\documentclass[sigconf]{acmart}
\AtBeginDocument{%
  \providecommand\BibTeX{{%
    \normalfont B\kern-0.5em{\scshape i\kern-0.25em b}\kern-0.8em\TeX}}}

\setcopyright{rightsretained} 
\copyrightyear{2021}
\acmYear{2021}
\setcopyright{rightsretained}
\acmConference[GECCO '21 Companion]{2021 Genetic and Evolutionary Computation Conference Companion}{July 10--14, 2021}{Lille, France}
\acmBooktitle{2021 Genetic and Evolutionary Computation Conference Companion (GECCO '21 Companion), July 10--14, 2021, Lille, France}\acmDOI{10.1145/3449726.3459431}
\acmISBN{978-1-4503-8351-6/21/07}

\begin{document}

\title{A Crossover That Matches Diverse Parents Together in Evolutionary Algorithms}

\author{Maciej~\'{S}wiechowski}
\email{m.swiechowski@mini.pw.edu.pl}
\orcid{0000-0002-8941-3199}
\affiliation{%
  \institution{QED Software}
  \city{Warsaw}
  \country{Poland}
}


\begin{abstract}
Crossover and mutation are the two main operators that lead to new solutions in evolutionary approaches. In this article, a new method of performing the crossover phase is presented. The problem of choice is evolutionary decision tree construction. The method aims at finding such individuals that together complement each other. Hence we say that they are diversely specialized. We propose the way of calculating the so-called complementary fitness. In several empirical experiments, we evaluate the efficacy of the method proposed in four variants and compare it to a fitness-rank-based approach. One variant emerges clearly as the best approach, whereas the remaining ones are below the baseline.
\end{abstract}


\maketitle

\section{Introduction}

Evolutionary techniques~\cite{simon2013evolutionary} have been serving as important tools in the Artificial Intelligence (AI) and Computational Intelligence (CI) tool-set.
In this article, we present a new method of selecting individuals for the crossover and grouping them into pairs that should undergo the crossover operation. This research has been inspired by an upcoming project aimed at evolving semantic-logical programs~\cite{logDL}. During initial research towards it, the closest problem of choice we found was the evolutionary decision tree induction. First of all, a decision tree can be structurally similar to a specific class of logical programs. Second of all, constructing a decision tree algorithmically is a well-understood problem, therefore we can focus on the analysis on the crossover phase. Evolving decision trees has been popular and dates back to before 2000~\cite{siegel1994competitively,papagelis2000ga}. The authors of~\cite{barros2013evolutionary} show that their method is capable of outperforming C4.5~\cite{c45}, which is a dedicated decision tree induction algorithm.

The authors of~\cite{dgea} state that diversity is one of the key factors in the performance of evolutionary algorithms. The diversity-guided algorithms are also subjects of~\cite{alam2012diversity} for Evolutionary Programming,~\cite{6277535} for Genetic Programming and \cite{7969516} for a Workforce Scheduling and Routing Problem. The means of measuring population diversity in genetic programming are summarized in~\cite{diversity}. In the broader field of evolutionary computation, there have been dedicated ways of measuring the diversity proposed, e.g.~\cite{4630928} for Particle Swarm Optimization (PSO) and ~\cite{nakamichi2004diversity} for Ant Colony Optimization (ACO).
 
\section{The Proposed Crossover Methods}

As a common part of the methods described below, except the \textit{Standard} one, we compute a measure, for each pair of individuals $A$ and $B$ in the population, which we will refer to as the \emph{complementary fitness}. It is a prediction of how fit they might potentially be combined. Afterwards, the pairs are sorted in descending order with respect to \emph{complementary fitness} and they perform the recombination until at least $K$ unique individuals has been recombined. $K = CR*N$, where $N$ is the population size and $CR$ is the crossover rate.\\

\noindent \textbf{\itshape Novel-2 Method: } the underpinning idea is to split the decision tree represented as in genetic programming into two parts and calculate the \emph{accuracy} for each part, respectively.
\begin{equation}
\label{eq:novel2}
\begin{aligned}
 \textrm{co-Fitness}_{AB} = \max(acc(A.LEFT), acc(B.LEFT)) \\
   					 + \max(acc(A.RIGHT), acc(B.RIGHT))
\end{aligned}
\end{equation}

\noindent \textbf{\itshape Novel-N Method:} let $Y^A_k$ denote the decision returned by the decision tree represented by the individual $A$ for the $k$-th sample in the training set. Let $Y^B_k$ be the decision for the $k$-th sample in case of the $B$ individual. Their \emph{complementary fitness} is calculated as the number of samples, in which either of the trees correctly predicted the decision:
\begin{equation}
\label{eq:cfN}
 \textrm{co-Fitness}_{AB} = \sum_{k=1}^{TS}[correct(Y^A_k,Y_k)  \vee correct(Y^B_k,Y_k)]
\end{equation}

\noindent \textbf{\itshape Standard Method:} we will use this name for the baseline. Here, the top $K$ fittest individuals perform the crossover.
Among this set, they are matched into pairs with uniform random probability. We have also experimented with a roulette-wheel sampling. It resulted in worse results for the considered problem.\\

\noindent \textbf{\itshape Hybrid-2 and Hybrid-N Methods} - in these methods, the population is sorted by the fitness value. Then, the first half of the parents for crossover are determined as in the \textit{Standard} method and the other half by \textit{Novel-2} or \textit{Novel-N} for \textit{Hybrid-2} and \textit{Hybrid-N}, respectively. 

\section{Results}

We have tested the five variants introduced in the previous section and compared them with each other. Each tested EA algorithm was initialized with a random population. For evaluation, we used \emph{the accuracy} (which is also the fitness function) of the best solution found so far by a respective method. Such a value was averaged over $400$ independent repeats of each experiment. In addition, we calculated the $95\%$ confidence intervals it. For the implementation of both EA and decision trees, we used the \emph{Grail AI library}~\cite{grail}.

The hybrid methods, i.e., \textit{Hybrid-2} and \textit{Hybrid-N} as well as the baseline  \textit{Standard} method outperformed the non-hybrid counterparts. Therefore, in Table~\ref{tab:results}, we show a summary of comparison among those three methods. A detailed explanation of the cause of this is one of our future plans. The possible explanation is that the pure novel methods do not facilitate enough population diversity.

\begin{figure}[!ht]
\includegraphics[width=\columnwidth]{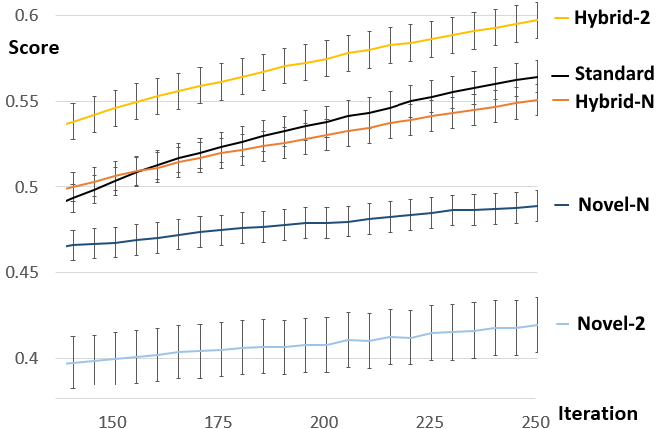}\
\caption{The scores obtained by each approach, plotted against EA's iteration for a smaller depth of the tree ($V=6$).}
\label{fig:fitness}
\end{figure}

In overall, the \textit{Hybrid-2} is the clear winner of the experiments. It achieved the highest $Score$ in all 8 experiments, however in 6 out of 8 experiments the advantage was statistically significant. The two experiments, in which it was not, were (1) with a lower crossover rate $CR=0.25$ and with (2) a smaller number of variables ($V=6$) in the decision tree. The first case suggests that the method gains advantage when having more individuals to work with. The second case suggests the problem needs to be complex enough for the method to show its advantage. For this problematic case, we present a plot of the best fitness value achieved by each method with respect to iteration in Figure~\ref{fig:fitness}. 
\begin{table}[ht]
\caption{A summary of results of the two best (in average) methods. The $1^{st}$ column contains parameters of the experiment: V - depth of the decision tree, N - the EA population size, TS - test set size, CR - crossover probability rate. In the $2^{nd}$ and $3^{rd}$ columns, we compare results of the two methods to the Standard method after 200 iterations of EA.}
\label{tab:results}
\begin{tabular}{lcc}
\hline
Experiment          & \multicolumn{1}{l}{Hybrid-2} & \multicolumn{1}{l}{Hybrid-N} \\
(V, N, TS, CR)      & \multicolumn{2}{c}{(compared to standard)}                  \\ \hline
(6, 200, 200, 0.5)  & better           & \textbf{worse}, significant     \\
(7, 200, 200, 0.5)  & \textbf{better}, significant  & similar, inconclusive                           \\
(8, 200, 200, 0.5)  & \textbf{better}, significant  & similar, inconclusive                              \\
(8, 200, 200, 0.25) & better               & similar, inconclusive                              \\
(8, 200, 200, 0.75) & \textbf{better}, significant  & \textbf{worse}, significant  \\
(8, 100, 200, 0.5)  & \textbf{better}, significant  & similar, inconclusive                              \\
(8, 400, 200, 0.5)  & \textbf{better}, significant  & similar, inconclusive                              \\
(8, 200, 400, 0.5)  & \textbf{better}, significant  & \textbf{worse}, significant  \\ \hline
\end{tabular}
\end{table}

\section{Conclusions}

In this paper, we have proposed a new method of choosing individuals for crossover. We evaluated the method using an evolutionary tree induction problem. The proposed method revolves around matching individuals into pairs, which have a high chance of producing fitter offspring. The method is suitable for scenarios in which the fitness is calculated as a sum (or aggregation, in general) of many parts and for each part a partial fitness value can be derived. We have shown that the best way to apply the proposed crossover procedure is by mixing it with the rank-based crossover selection. Such a merger is stronger than either of the methods alone what has been confirmed in 8 empirical experiments.

\bibliographystyle{ACM-Reference-Format}
\bibliography{main}

\end{document}